\documentclass[letterpaper, 10 pt, conference]{ieeeconf}

\usepackage{tabularx,booktabs}
\usepackage[flushleft]{threeparttable}
\usepackage{nameref}
\usepackage{amsmath}
\usepackage[utf8]{inputenc}
\usepackage{booktabs}
\usepackage{cite}
\usepackage{graphicx}
\usepackage{color}
\usepackage{url}
\usepackage{cite}
\usepackage{comment}
\usepackage{amsfonts}
\usepackage[per-mode=symbol]{siunitx}
\pdfoutput=1

\usepackage{hologo}
\DeclareSIUnit{\fps}{ fps }

\IEEEoverridecommandlockouts          
\begin{document}

\title{\LARGE \bf System Configuration and Navigation of a Guide Dog Robot:\\Toward Animal Guide Dog-Level Guiding Work}

\author{Hochul Hwang$^{\dagger}$, Tim Xia$^{\dagger}$, Ibrahima Keita, Ken Suzuki, Joydeep Biswas$^{1}$, Sunghoon I. Lee$^{*}$, and Donghyun Kim$^{2}$
\thanks{$^{\dagger}$ Authors contributed equally, $^{*}$ Co-corresponding author}
\thanks{$^{1}$ University of Texas at Austin {\tt\small joydeepb@cs.utexas.edu}}%
\thanks{$^{2}$University of Massachusetts Amherst, 140 Governors Dr, U.S. {\tt\small donghyunkim@cs.umass.edu}}
}

\maketitle
\thispagestyle{empty}
\pagestyle{empty}

%%%%%%%%%%%%%%%%%%%%%%%%%%%%%%%%%%%%%%%%%%%%%%%%%%%%%%%%%%%%%%%%%%%%%%%%%%%%%%%%
\begin{abstract}
    A robot guide dog has compelling advantages over animal guide dogs for its cost-effectiveness, potential for mass production, and low maintenance burden. However, despite the long history of guide dog robot research, previous studies were conducted with little or no consideration of how the guide dog handler and the guide dog work as a team for navigation. To develop a robotic guiding system that is genuinely beneficial to blind or visually impaired individuals, we performed qualitative research, including interviews with guide dog handlers and trainers and first-hand blindfold walking experiences with various guide dogs. Grounded on the facts learned from vivid experience and interviews, we build a collaborative indoor navigation scheme for a guide dog robot that includes preferred features such as speed and directional control. For collaborative navigation, we propose a semantic-aware local path planner that enables safe and efficient guiding work by utilizing semantic information about the environment and considering the handler's position and directional cues to determine the collision-free path. We evaluate our integrated robotic system by testing guide blindfold walking in indoor settings and demonstrate guide dog-like navigation behavior by avoiding obstacles at typical gait speed (\SI{0.7}{\meter\per\second}). 
\end{abstract}
%%%%%%%%%%%%%%%%%%%%%%%%%%%%%%%%%%%%%%%%%%%%%%%%%%%%%%%%%%%%%%%%%%%%%%%%%%%%%%%%

\section{Introduction}
According to the World Health Organisation (WHO), more than 2.2 billion people in the world have near or distance vision impairments~\cite{WHO}. Guide dogs (GD) have been considered as an effective way to improve the mobility and independence of visually impaired individuals~\cite{miner2001experience, whitmarsh2005benefits, wiggett2008experience}.
However, only about 2\% of blind people in the U.S. work with guide dogs~\cite{howmany_guidedog}, and the percentage is expected to decrease worldwide mainly due to the lack of supply. Fundamental challenges of expanding the supply are the significant cost and time that are required to train, deploy, and maintain animal guide dogs. The cost for training one guide dog is over \$50,000 USD and the time for breeding, raising, training, and deploying takes up to two years. Guide dogs' work span is typically less than ten years and they require lifelong care from their handler, including medical care, feeding, and daily walks. As a result, they are not suitable for people who cannot raise a dog due to various reasons, such as financial affordability, allergies, and physical limitations (e.g., due to motor impairments or natural aging)~\cite{miner2001experience, wiggett2008experience, hersh2010robotic}. To remedy the issues and embrace the potential of mass production, a cost-effective and sustainable guide dog robot could be a significant contribution to the visually impaired population.

\begin{figure}
    \centering
    \includegraphics[width=0.9\columnwidth]{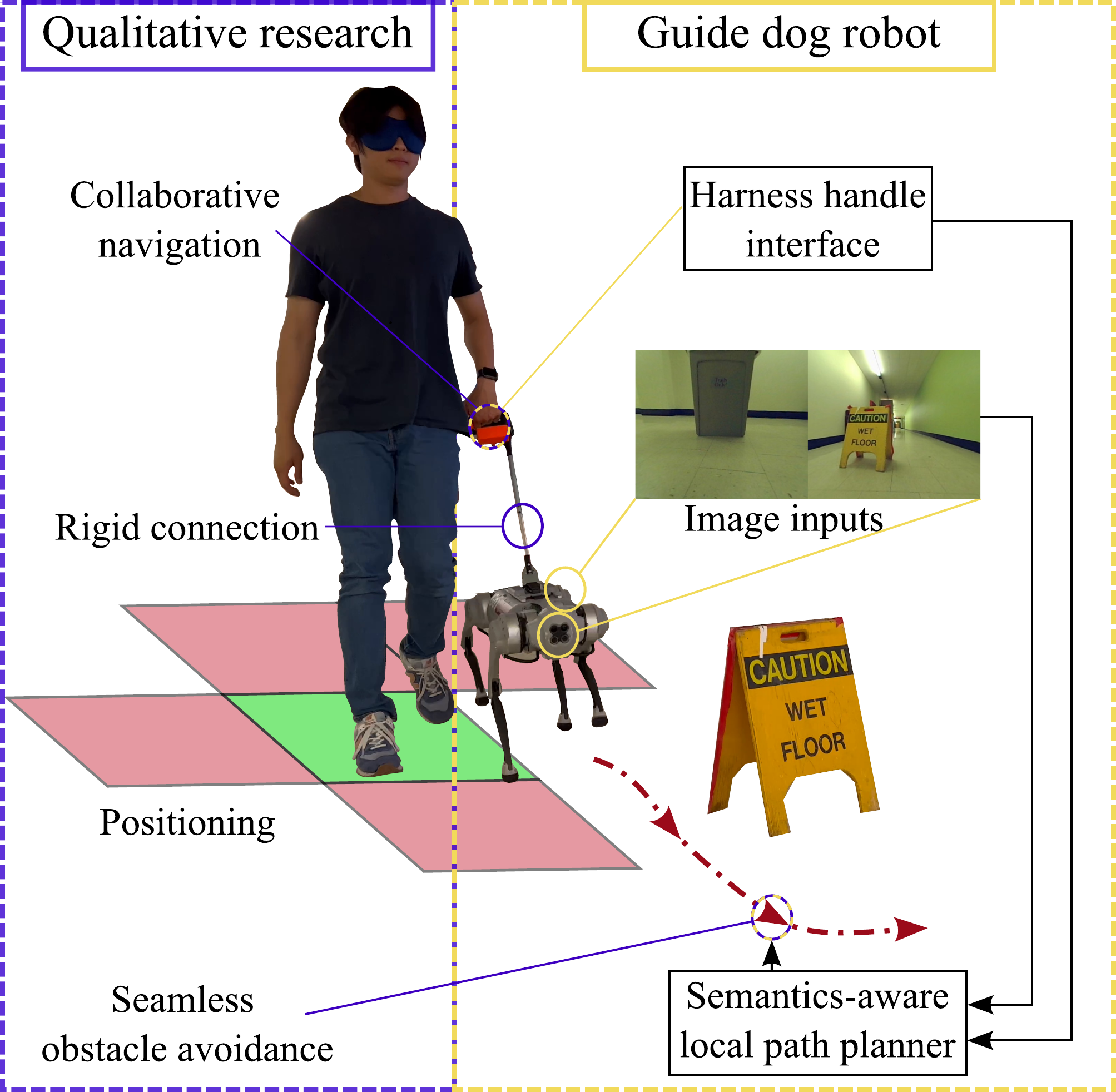}
    \caption{Development of a guide dog robot based on qualitative research. Summer collaboratively works with the handler by reflecting the handler's directional suggestions to reach the destination in a safe and efficient manner by seamlessly avoiding obstacles similar to an animal guide dog.}
  \label{fig:scheme}
\end{figure}

This strong need inspired researchers, and various navigation-assistant devices have been developed for a long time. The early study by \cite{Tachi} evidenced how a robotic system can support the navigation works of a person with visual impairment. Multiple wheeled mobile robots have been developed from 1976 to 1983 and demonstrated autonomous navigation and obstacle avoidance while guiding a person~\cite{meldog}. On the other hand, researchers also developed various forms of assistive devices such as smart canes~\cite{smartcane, slade2021multimodal}, vibrating belts~\cite{belt}, hand-held camera traversability estimators~\cite{GOnet}, and head mount devices \cite{headmount}. Throughout the research, the perception assistant has been actively studied, but the mobility of the assistant device was less highlighted. The recent progress in locomotion control of quadruped robots opened the chance to utilize the legged robots for guide dog robots, and several studies using a quadruped robot have been reported~\cite{xiao2021robotic, chen2022quadruped, hamed2019hierarchical}. However, we discover that the existing guide dog robots do not adequately address how the handlers interact with their animal guide dogs in the real world, which has been established over several decades to support comfortable mobility and navigation efficiency.

For example, having a rigid connection is crucial for the handlers to perceive the motion of the guide dog and navigate in a safe manner~\cite{guerreiro2019cabot}. In addition, positioning the handler right next to the robot is important to navigate through sharp turns and narrow paths. However, \cite{xiao2021robotic, chen2022quadruped} used a soft leash and \cite{hamed2019hierarchical} located a person behind a robot. Moreover, previous studies have mainly focused on robot control under the handler's pulling force or autonomous navigation rather than interactive communication during a guiding work of a handler-guide dog unit. Besides the handle design and the handler's position, various aspects should be considered in the guiding system design. Rich literature about guide dog work and training methods exists \cite{michael84, wiener2010foundations, Christie20}, but less guidelines have been disclosed for even basic robot specifications such as size, weight, battery life, sensors, walking speed, required robustness of walking controllers, and so on \cite{hersh2010robotic}. For the navigation algorithm development, we have to precisely understand the roles of a handler and a dog because guiding work is different from autonomous driving, which brings a person to the target location without intermittent direction commands from the driver. 

In this study, we aim to identify important design components of a guide dog robot to support effective navigation and obstacle avoidance based on the established interaction mechanisms between the handler and animal guide dogs. To understand the interaction mechanisms, we conducted interviews with guide dog handlers and trainers, followed by several observation sessions to investigate how the handler navigates with the guide dog in real-world settings. Our team also performed blindfold walking with four different guide dogs under a guide dog trainer's supervision to understand the navigation process better. Based on qualitative data analysis, we configure the development directions of a guide dog robot and build an interactive navigation scheme and local path planning algorithm.

In terms of hardware design, we adopted the handle developed for animal guide dog users and installed a custom-designed button interface to allow the handlers to adjust speed and give walking direction cues, which are the most frequently used cues in guiding work. The handle can be completely detachable from the robot's body to help the handlers to easily keep the robot underneath their seat when they use public transportation. For the local path planner, we primarily focus on rapid response to the environment change to safely navigate a dynamic environment. Special effort is made to enable seamless pedestrian avoidance while maintaining the handler's normal walking speed to provide a comfortable guiding work experience. 

The proposed semantics-aware local path planner selects a path from a cost map made with a pretrained semantic segmentation network~\cite{MITseg}  without extra geometrical route computation to save computation power. In addition, our method relies only on built-in RGB cameras without depth estimation and a standard NVIDIA GPU computing unit. Similar navigation approaches that use pretrained image processing backbones \cite{CognitiveMapping, ZhuTarget} have been proposed, but less have been explored to directly utilize semantic segmentation as a local path planner. \cite{KimSemantic} proposed to utilize off-the-shelf semantic segmentation for indoor mobile robots but did not deploy the methods on actual robotic systems. \cite{Yeboah} explored the idea of indoor navigation based on semantic segmentation, in which they trained various segmentation networks and analyzed their computational efficiency, and in \cite{RyusukeIndoor, MathJournal}, the researchers deployed semantic segmentation on an actual robot for indoor navigation, and both relied on the egocentric view of the segmented scene for target point creation and subsequent target point following. However, such methods are sensitive to the noise of the segmentation network, which is inappropriate for safety-critical applications. Our planner provides robustness against the segmentation error by employing simple cost map-based path selection and rule-based stop criteria.

Primary contributions of our work are summarized as follows: 1) human-centered system integration of a guide dog robot based on qualitative research with guide dog handlers and trainers, 2) development of a guiding work framework with a safety-oriented local path planner that imitates animal guide dogs' navigation works, and 3) experimental evaluation of the robot in real-world indoor environments and demonstration of animal guide dog-like behavior at a preferred walking speed. The video 

\begin{table}
  \caption{Participant demographics}
  \label{tab:qual}
  \setlength{\tabcolsep}{0.7\tabcolsep}
  \centering
  \begin{tabular}{ *{5}{c} }
    \toprule
    \textbf{Subject} & \textbf{Age} & \textbf{Gender} & \textbf{Experience (years)} & \textbf{Session type} \\
    \midrule
    H01 & 60 & F & 36 & Interview, observation \\ % I01, I02 (observation), O01
    H02 & 69 & M & 30 & Interview, observation \\
    H03 & 63 & M & 11 & Interview, observation\\
    H04 & 69 & F & 30 & Interview, observation\\
    H05 & 59 & M & 21 & Interview, observation\\
    T01 & 65 & M & 45 & Interview \\
    T02 & 52 & M & 19 & Interview, observation\\
    T03 & 58 & M & 35 & Interview, observation\\
    \bottomrule
  \end{tabular}
  \vspace{-3mm}
\end{table}

\section{Guide dog robot configuration}
Guide dogs work with the handler as a team, called a \textit{unit}, to safely navigate to the desired destination. Also, unlike a white cane, guide dogs provide efficient navigation by seamlessly avoiding obstacles at a walking speed. Based on qualitative research with five guide dog handlers (GDHs) and three guide dog trainers (GDTs), we discovered 1) core components that must be considered when developing a guide dog robot to ensure safety and 2) advanced components to enhance efficiency in navigation and increase comfort for the handler. We elaborate on our guide dog robot’s configuration by specifying the hardware requirements for real-world deployment. The participants' demographics are summarized in Table~\ref{tab:qual}.

\begin{figure}
    \includegraphics[width=0.95\columnwidth]{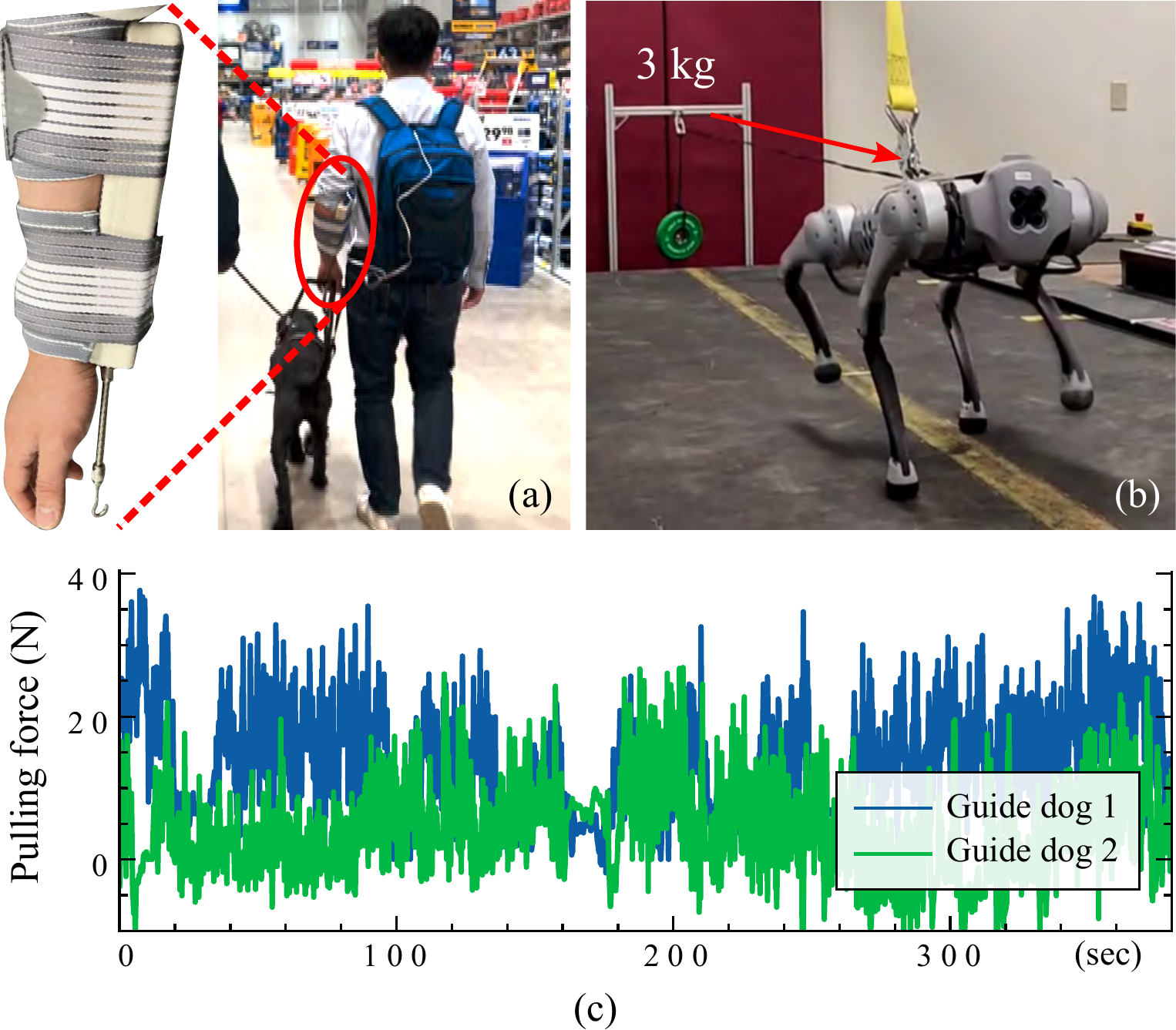}
    \caption{Pulling force measurement of animal guide dogs and the guide dog robot, Summer. (a) We performed blindfold walk with different guide dogs. A digital force gauge was used to measure the pulling force while walking. (b) Go1's built-in controller is evaluated on a testbed that pulls the robot with the horizontal force up to \SI{34.32}{\newton} (\SI{3.5}{\kilogram}). (c) Pulling forces measured during blindfold walking with two different guide dogs.}
    % Summer's maximum pulling force is also plotted for comparison.
  \label{fig:pulling}
  \vspace{-3mm}
\end{figure}

\subsection{Hardware requirements to serve as a guide dog robot}
To address the various physical attributes of various blind or visually impaired individuals, guide dog trainers walk with the potential guide dog handler beforehand and match with the appropriate guide dog considering features including the size and pulling force. For a guide dog robot, compact size and small weight are desirable as they allow handlers to be able to independently carry and put the robot under seats, especially when they use public transportation such as buses, trains, or taxi cabs. For an animal dog, this can be easier as they have a flexible body as one of the handlers mentioned:

\begin{quote}
\textit{“Whereas a guide dog, being squishy, you train him to, even a dog his size, get on the floor area and lay down.”} - \textbf{H01}
\end{quote} 

Although minimizing the weight is beneficial for portability, it should still be heavier than a certain mass to handle the nominal pulling force of a handler and to address emergent stops. To identify the range of pulling force in guide dog work, we measured the pulling force with a force gauge during our blindfold walking with different dogs as in Fig.~\ref{fig:pulling}' (a). We found that the pulling force of a guide dog remained mostly under $40 \si{\newton}$ even in the case of a dog with strong pulling strength (guide dog 1 in Fig.~\ref{fig:pulling} (c)). We experimentally validated that the robot's locomotion controller is robust enough to manage a similar external disturbance by letting the robot walk while being pulled by \SI{3}{\kilogram} weight as shown in Fig.~\ref{fig:pulling} (b). 

\subsection{Core components in GD work to ensure safety}
\subsubsection{Rigid connection and correct positioning}
\begin{quote}
\textit{“I think it can't be done without some sort of device that you have some sort of handle. … Because, when you're working with whatever to get someplace, you are feeling that thing move. … if I'm holding a dog or holding a two-point handle to something if it stops, I would stop.”} - \textbf{H01}
\end{quote}

As one of the GDH subjects mentioned, in guide dog work, the rigid connection between the handle and the guide dog is critical. Most guide dog handlers primarily use a harness while walking since it can provide instant feedback of the guide dog’s motion while it is challenging to interpret the motion of the guide dog using only a soft leash. 

Walking at a proper position relative to the guide dog (next to the dog's hindquarters) while walking is also critical for various safety purposes. For example, when the animal guide dog stops at a curb, the handler can use the foot within a single step to reach out to recognize and locate the curb. Also, it is ideal to be in the recommended position to reduce the range of motion and turning speed for safety. Additionally, note that most guide dogs are trained to walk on the handler's left-hand side because most handlers are right-handed. This enforces the handler to use their left hand to hold the harness handle and to use their right hand to perform additional tasks or search around when a dog stops. This setup can be uncomfortable for left-handed handlers, but it is hard to accommodate them because switching the position requires additional training.

Considering the aforementioned safety concerns, we use a rigid harness handle structure provided by Ruffwear. The length-adjustable design is expected to support various potential handlers, having different physical features, to maintain the proper position while walking. On top of the safety benefits, the rotating handle structure enables easy transition between left-handed and right-handed handlers. Also, the detachable harness structure is desirable considering the handler's portability.

\begin{figure}
    \centering
    \includegraphics[width=\columnwidth]{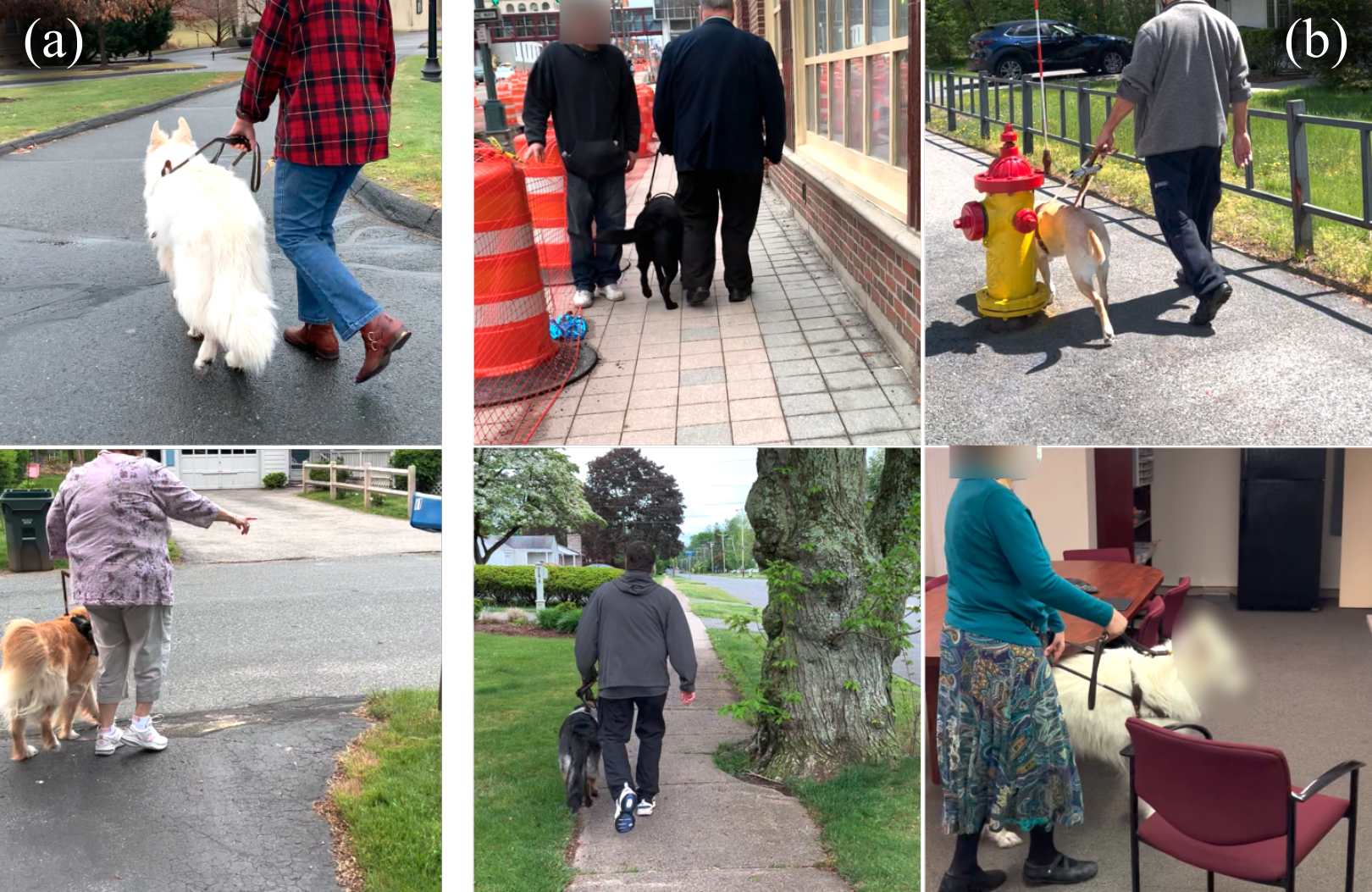}
    \caption{Observation of the guide dog handler's walking with their guide dogs. (a) The handler-guide dog unit is turning based on the handler's directional cue. (b) The handler-guide dog units are seamlessly avoiding obstacles (e.g., a chair, a tree, and a person) in their local community.}
  \label{fig:qual}
\end{figure}

\subsubsection{Collaborative navigation}
\label{sec:collaborative_navigation}
Unlike autonomous navigation, in most guiding work, the handler needs to provide cues to the guide dog to some extent, especially when initiating turns and deciding when to cross a street, which can also be observed in Fig~\ref{fig:qual} (a). In the basic setup between a handler and a guide dog, the handler is responsible for knowing the directions to the target location based on cognitive maps and making decisions about whether it is safe to make a move by interpreting nonvisual information~\cite{wiener2010foundations, giudice2008blind}. 

\begin{quote}
\textit{“We have to have some idea of where you want to go. So it becomes the person's job to tell them straight, forward, left, right, find the way, look for whatever they're aiming for.”} - \textbf{T03}
\end{quote} 

We implement such collaborative navigation by 1) designing a button interface that takes the combination of touch sensor signals provided by the handler as input and 2) allowing our proposed semantics-aware local path planner to take this input into the path selection process. With such implementation, Summer interprets the handler’s directional suggestions to safely take turns when available. 

Based on the feedback obtained from the GDHs, we designed a wireless button interface that allows the handler to suggest directional cues to Summer. We minimized the number of buttons to reduce the complexity of usage. These two buttons are digitally connected to an Arduino Nano which is attached to a Bluetooth HC-05 module. Note that the wireless communication paradigm allows the harness handle to maintain its detachability. With a single press of the button, a handler can give a directional command to the local path planner that takes the command into consideration when deciding the most desirable path which the details are provided in Section~\ref{sec:navigation}. 

\subsection{Advanced components in GD work to enhance efficiency and comfort}
\subsubsection{Seamless obstacle avoidance at high walking speed}

Guide dogs not only ensure the safety of the handler by avoiding collision but they allow higher travel speed by seamlessly maneuvering around obstacles that we found from interviews and the observation session (Fig~\ref{fig:qual}(b)). This smooth mobility is difficult to get with a white cane because a user needs to find a way to avoid obstacles in a zig-zag manner based on tactile information~\cite{dos2021electronic}. One of the guide dog trainers gives a clear idea of how seamlessly an animal guide dog can avoid obstacles. 

\begin{quote}
\textit{“Little bumps in the road and you'll realize that sometimes if there's a bunch of obstacles, whether it's traffic cones, trees, fire hydrants, whatever it is, you might never know them there. You can have a guide dog for 10 years and you'll go, `Oh I didn't know that was there,' because the dog just fluidly goes around. It's a little different than using your cane where your cane you're aware of all the obstacles.”} - \textbf{T03}
\end{quote} 

Here, we focus on developing a local path planner to provide smooth obstacle avoidance in guiding work without slowing down or stopping the handler unless they are necessary as explained in Section~\ref{sec:navigation_details}. We also allow the handler can control the walking speed using the button interface because the walking speed varies by handlers; the dog's walking speed is one of the most important parameters to matching a proper handler-guide dog unit. 

\begin{quote}
    \textit{"Fidelco (training school), before they give you a dog, will match up your walking speed with a harness that they pull around to see how fast, slow, medium, or fast."}- \textbf{H03}
\end{quote}

\begin{figure}
    \centering
    \includegraphics[width=\columnwidth]{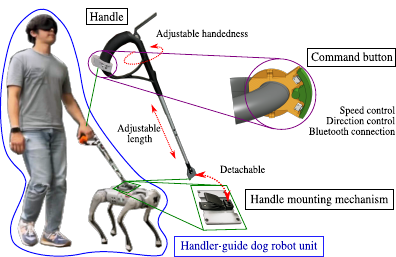}
    \caption{Handle interface design that features speed/direction control button, adjustable handle, and detachable hook for portability.}
  \label{fig:button}
\end{figure}

The walking speed can be controlled incrementally by pressing the button consecutively. As an example, the walking speed of Summer will increase by \SI{0.05}{\meter\per\second} if the upper button is pressed twice as shown in Fig.~\ref{fig:button}.

\begin{figure*}
    \centering
    \includegraphics[width=\textwidth]{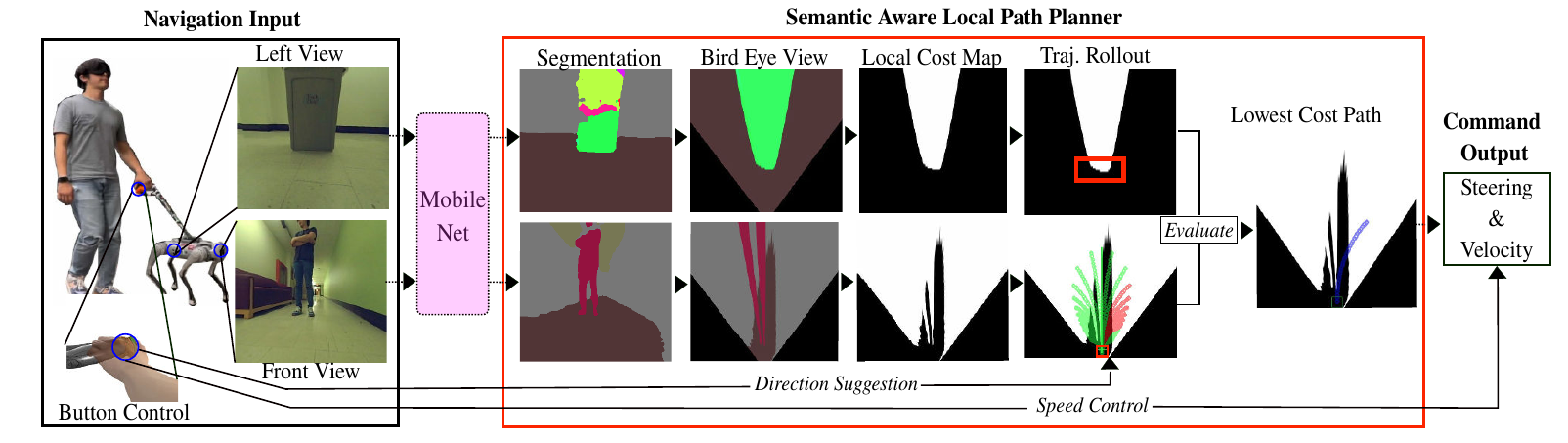}
    \caption{Navigation pipeline. Images from front and side camera are passed into a segmentation network. The segmented images are then projected down to bird-eye-view and subsequently converted into a binary traversable and non-traversable local cost map. Trajectories are rolled out on this cost map. The button interface is used to bias trajectory selection. When the best path is selected, the steering command is fed into the built-in controller. The button interface can also be used to adjust the forward velocity of the controller. Velocity is also adjusted based on cost boxes in the front and the left local cost maps in case of emergencies for obstacle clearing.}
  \label{fig:scheme-2}
  \vspace{-3mm}
\end{figure*}

\section{Guiding work framework}
\label{sec:navigation}

As explained in Section.~\ref{sec:collaborative_navigation}, a handler provides orientation cues such as forward, left, or right and a guide dog take care of local path planning by safely and seamlessly avoid collision or other risk for the handler. Based on the standard guiding work setup, we focused on developing a local path planner that can provide rapid response to the environment change, avoiding various obstacles including walking pedestrians without incurring uncomfortable jerky movements or sudden stops unless they are necessary. One challenge was avoiding walking pedestrians coming from opposite direction while maintaining the handler's walking speed. Considering the average speed of humans ($\approx 1~\si{\meter\per\second}$) and perception camera sight ($ 2 - 3 ~\si{\meter}$), the robot has around only $1~\si{\second}$ before it hit the person. Therefore, we targeted to maximize the update frequency by removing computationally complex process and by minimizing sensor inputs. Our image segmentation-based local planner plans and evaluates paths at a rate of 20 Hz and is capable of robust static and dynamic obstacle avoidance. Here we present the main components of our navigation pipeline. 

\subsection{Local path planner}
We assume the robot is constantly moving forward, and employ a local planner based on a trajectory roll-out algorithm that samples 40 trajectories parameterized by minimum and maximum curvature. The 40 curvatures are rolled out in front of Summer and each is then individually evaluated according to length and visual cost. At each time step, the best path is selected and Summer adjusts its steering angle according to the curvature of the selected best path.

In order to evaluate the visual cost associated with the rolled-out trajectories, a visual cost map is built based on the semantic categories of the objects in the Summer's field of view. We use the front- and side-facing camera equipped on Summer as our main input. The camera operates at the rate of $30~\si{\fps}$, and two  $464\times 400$ sized RGB images are streamed from Summer to a laptop in the form of ROS messages \cite{ros}. Once the image is received by the laptop, it is processed by a pretrained network with a MobileV2Dialted encoder \cite{Mobile} and a convolutional decoder and converted into two $464\times 400$ sized matrix, where pixel values are converted into one of 150 semantic classes. As we are only concerned about the traversable class, \textit{floor}, all 150 classes are converted into binary classes, i.e., $\text{\textit{floor}} = 0$ vs. $\text{\textit{non-floor}} = 1$. The resulting binary $464\times 400$ matrices are then scaled by 200 and converted into a grayscale image. The grayscale image is then mapped down to BEV using a homography matrix and is used as a cost map to evaluate trajectories. 

\begin{figure*}
    \centering
    \includegraphics[width=2\columnwidth]{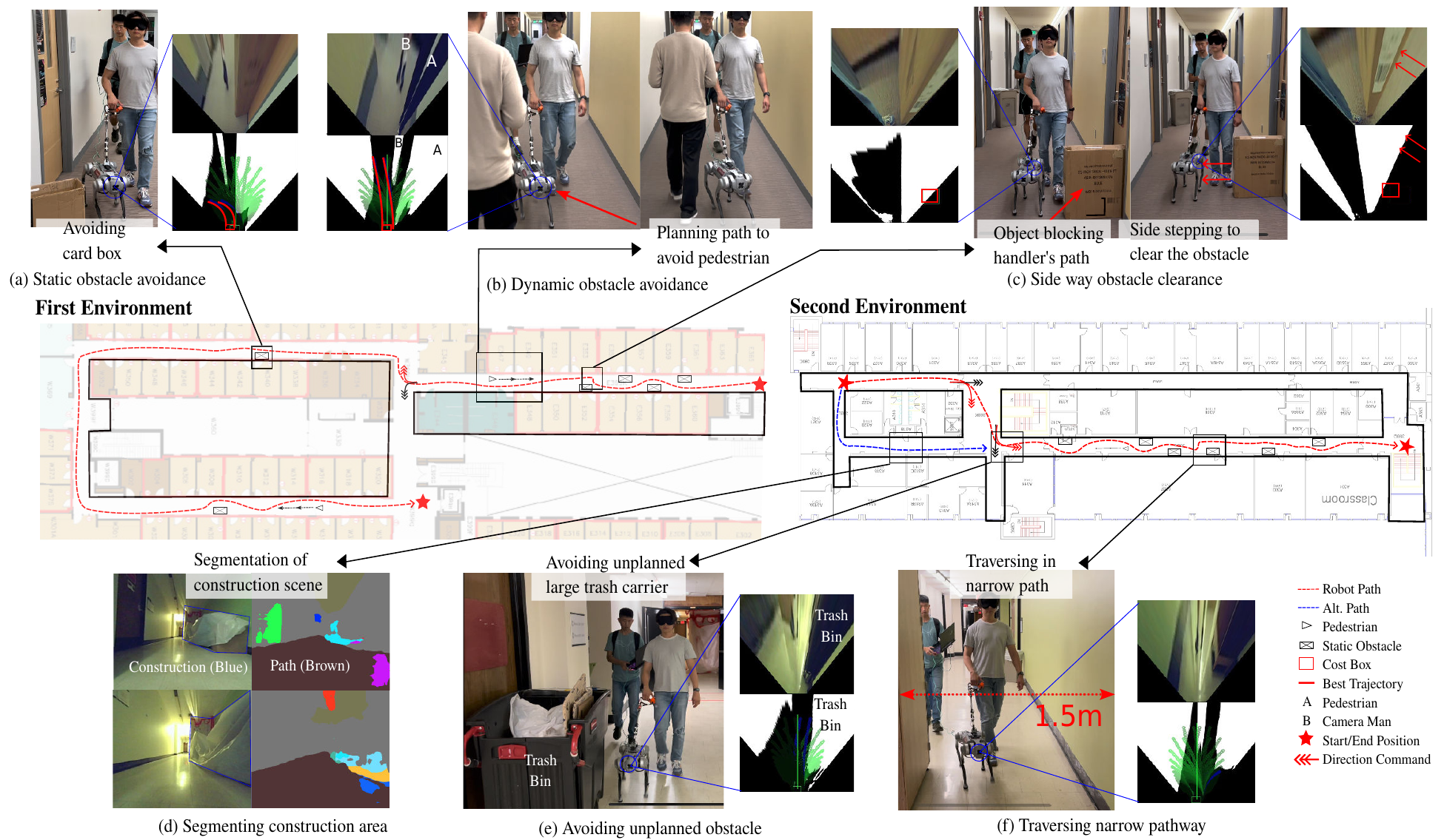}
    \caption{{\bf Indoor guiding work experiments} (a) The bird-eye view and the cost map show how the robot recognize the environment and convert the view to the cost map. Among the 40 path candidates (green), the path with minimal cost is selected (red). (b) The right image shows how the pedestrian is represented in the cost map. (c) Summer uses a side-view camera to secure the enough traversable space for the handler. When upcoming obstacles are detected, which is shown as the white area under the red cost box in the cost map, Summer makes side stepping until the obstacle is cleared. (d) (e) Our method can find a path as long as the side walk is properly detected. Therefore, guiding work is robust to the situations where unfamiliar objects exist (e.g., construction site) (f) Summer can navigate through narrow path with $1.5~\si{\meter}$ width. }
  \label{fig:obstacle}
  \vspace{-3mm}
\end{figure*}

\subsection{Trajectory adjustment, emergency stop, and path clearance}
\label{sec:navigation_details}
We augmented the trajectory roll-out algorithm by simultaneously rolling out parallel trajectories, thus evaluating an extended trajectory space that takes into account Summer's body width as well as the handler's position. Furthermore, we implemented an emergency stop mechanism in the form of a visual cost box located directly in front of Summer in it's BEV local cost map. If an obstacle comes too close into proximity or if navigation itself chose the wrong path and was heading straight into an obstacle, then the segmented object or non-traversable path would enter the visual cost box, and based on a cost threshold, the forward velocity command will halt, preventing Summer and the handler from crashing into the obstacle. An additional cost box is instantiated in Summer's left BEV local cost map. The logic of the mechanism is identical, however, this box's function is to instruct Summer to step to the right whenever there is not enough space on it's left side for the handler to traverse. 

\subsection{Handler's direction and speed control}
When the up button attached to the handle is pressed, a right turn command is issued, and this influences the trajectory evaluation algorithm by reducing the costs assigned to the paths that lead to the right side, thereby biasing Summer to turn right. The same mechanism applies to the left when the down button is pressed. Moreover, we implemented a speed control mechanism, where if the up button is pressed twice, Summer's forward velocity increases, enabling the guided unit to traverse at a faster pace. The same mechanism applies to slowing down if the down button is pressed twice. 

\section{Experimental Results}
To demonstrate the robustness of the proposed local path planner, we performed blindfold guiding work in two different indoor environments, which differ in basic layout, flooring, and wall texture. In the first environment, the total length of the navigation task is $105~\si{\meter}$, and in the second environment, the total length of the navigation task is $90~\si{\meter}$. In both environments, the task is to traverse through a hallway while avoiding collision with static and dynamic obstacles. Obstacles included in the experiments are card boxes, blue mats, trash cans, a wet sign, and a moving person. Both environments contain turns where the handler needs to issue directional commands and a long hallway where the path is cluttered with obstacles as close as $2~\si{\meter}$ apart. During the experiment, a sighted person is blindfolded and acts as a handler. The handler is situated on the left side of Summer to demonstrate that the handle can be easily adjusted for handedness preferences. An operator will be following the handler while carrying a GPU-containing laptop. In real guiding work, a handler gives direction cues, but in our experiment, the operator provides signals on when to give directional cues to the handler because the blindfolded handler is not well accustomed to the layout of the test environments. 

\subsection{Obstacle Avoidance and Command Following}
In both environments, Summer and the handler successfully complete the task without a third person's intervention except for the operator's directional cues. During the tests, Summer smoothly walks around obstacles and keeps itself and the handler from collision. In cases where the turn is narrow, Summer stops and relies on its left visual cost map to adjust its position such that the handler has enough traversable space. In walking pedestrian avoidance, Summer leverages its high-frequency planner and responsibly swerves around the incoming person. Because the direction command does not override but only biases the path selection algorithm, the timing of the command does not need to be precise. In our experiments, as long as the blindfolded handler issued a direction cue before the turn, Summer was able to follow the direction issued by the handler. 

 \subsection{Robust Behavior in Unplanned Scenarios}
Several unplanned scenarios occurred during the experimentation process. Construction took place in one experimental condition that drastically changed the appearance of the hallway and reduced the width of the traversable area. Despite the appearance change, the reduction in traversable area, vaguely identified objects, Summer finds a traversable path as long as the side walk is properly detected, which is true most cases. Another unplanned scenario occurred when a large trash carrier was placed on the planned route, blocking a significant portion of the traversable area. Despite this, Summer was able to successfully navigate around the obstacle. 

\section{Conclusions}
In this letter, we present a safe and resource-light guide dog robot that is designed based on insights gathered from qualitative research. We utilize two built-in RGB cameras on the Go1 robot and an off-the-shelf semantic segmentation network to establish a local path planner. We implement navigational mechanisms that allows the robot to stop during emergencies and avoid obstacles while accounting for the robot's body and position of the human handler. Custom harness handle interface is designed and fabricated for the interactive guiding work. Finally, we demonstrate the robustness of our method in two different indoor environments.

\section*{ACKNOWLEDGMENT}
The study materials used in this study has been reviewed and approved by the University of Massachusetts Amherst IRB, Federal Wide Assurance \# 00003909 (protocol ID: 3124). We acknowledge Unitree Robotics for providing a Go1 robot featured with low-level controller accessible software and for technical support to enable this study. We also acknowledge Ruffwear for providing the initial harness design suitable for our application.


%%%%%%%%%%%%%%%%%%%%%%%%%%%%%%%%%%%%%%%%%%%%%%%%%%%%%%%%%%%%%%%%%%%%%%%%%%%%%%%%

%%%%%%%%%%%%%%%%%%%%%%%%%%%%%%%%%%%%%%%%%%%%%%%%%%%%%%%%%%%%%%%%%%%%%%%%%%%%%%%%

%%%%%%%%%%%%%%%%%%%%%%%%%%%%%%%%%%%%%%%%%%%%%%%%%%%%%%%%%%%%%%%%%%%%%%%%%%%%%%%%
% \section*{APPENDIX. Interview Questions}
% \label{sec:appendix}

%%%%%%%%%%%%%%%%%%%%%%%%%%%%%%%%%%%%%%%%%%%%%%%%%%%%%%%%%%%%%%%%%%%%%%%%%%%%%%%%

\bibliographystyle{./IEEEtran} % use IEEEtran.bst style
\bibliography{./IEEEabrv,./reference}

\end{document}